\documentclass[10pt,conference, letterpaper]{IEEEconf}
\newcommand{\bfa}{{\bm{a}}}

\newcommand{\bfb}{\mathbf{b}}

\newcommand{\bff}{\bm{f}}
\newcommand{\hbff}{\hat{\bm{f}}}

\newcommand{\bfF}{\bm{F}}

\newcommand{\bfG}{\bm{G}}

\newcommand{\bfg}{\bm{g}}
\newcommand{\bfh}{\bm{h}}
\newcommand{\bfH}{\bm{H}}

\newcommand{\bfI}{\bm{I}}

\newcommand{\bfp}{\bm{p}}
\newcommand{\hbfp}{\hat{\bm{p}}}

\newcommand{\bfq}{\bm{q}}
\newcommand{\hbfq}{\hat{\bm{q}}}

\newcommand{\bfR}{\bm{R}}

\newcommand{\hbfR}{\hat{\bfR}}

\newcommand{\bfS}{\bm{S}}

\newcommand{\bfv}{\bm{v}}

\newcommand{\hbfv}{\hat{\bfv}}

\newcommand{\bfW}{\bm{W}}
\newcommand{\bfw}{\bm{w}}

\newcommand{\bfx}{{\bm{x}}}

\newcommand{\hbfx}{\hat{\bm{x}}}

\newcommand{\bfy}{\bm{y}}

\newcommand{\boeta}{\bm{\eta}}

\newcommand{\boomega}{\bm{\omega}}

\newcommand{\hboomega}{\hat{\boomega}}

\newcommand{\boPhi}{\bm{\Phi}}

\newcommand{\bosigma}{\bm{\sigma}}

\newcommand{\bovartheta}{\bm{\vartheta}}

\newcommand{\bfzero}{\bm{0}}

\newcommand{\covQ}{\bm{\mathcal{Q}}}
\newcommand{\covR}{\bm{\mathcal{R}}}

\newcommand{\covP}{\bm{\mathcal{P}}}

\newcommand{\diag}{\mathrm{diag}}

\newcommand{\ned}{\{$n$\}}
\newcommand{\ecef}{\{$e$\}}
\newcommand{\body}{\{$b$\}}
\newcommand{\inertial}{\{$i$\}}
\newcommand{\geo}{\{$g$\}}

\usepackage{graphicx} 
\usepackage[english]{babel}
\usepackage[utf8]{inputenc}
\usepackage{csquotes}

\usepackage[hidelinks]{hyperref}

\usepackage{amsmath}
\usepackage{amssymb}
\usepackage[per-mode = symbol]{siunitx}
\usepackage{xfrac}
\usepackage{bm}

\usepackage{tabularx}
\newcolumntype{Y}{>{\hsize=4\hsize}X}
\usepackage{booktabs}

\usepackage[capitalize]{cleveref}
\crefname{appendix}{App.}{Apps.}
\crefrangeformat{figure}{Figs.~#3#1#4--#5#2#6}
\crefrangeformat{table}{Tabs.~#3#1#4--#5#2#6}
\crefformat{equation}{(#2#1#3)}
\crefrangeformat{equation}{(#3#1#4)-(#5#2#6)}

\usepackage{todonotes}

\IEEEoverridecommandlockouts
\usepackage{cite}
\usepackage{amsmath,amssymb,amsfonts}
\usepackage{algorithmic}
\usepackage{graphicx}
\usepackage{textcomp}
\usepackage{siunitx}
\usepackage{caption}
\usepackage{subcaption} 
\usepackage{xcolor}
\usepackage[nolist]{acronym}

\usepackage{tikz}
\usetikzlibrary{positioning,fit}
\usetikzlibrary{calc}
\usetikzlibrary{arrows.meta}
\usetikzlibrary{tikzmark} 
\usepackage{pgfplots}
\usepackage{wasysym}

\DeclareSIUnit{\rad}{rad}
\DeclareSIUnit{\nanotesla}{nT}
\DeclareSIUnit{\hertz}{Hz}
\DeclareSIUnit{\kmh}{km/h}

\def\BibTeX{{\rm B\kern-.05em{\sc i\kern-.025em b}\kern-.08em
    T\kern-.1667em\lower.7ex\hbox{E}\kern-.125emX}}
\pgfplotsset{compat=1.18}

\usepackage{todonotes}

\usepackage{fancyhdr}
\fancypagestyle{preprintnotice}{
  \fancyhf{} 
  \fancyhead[C]{\small\itshape This work has been submitted and accepted to the IEEE for publication. \\ 
  Copyright may be transferred without notice, after which this version may no longer be accessible.}
}

\begin{document}
\begin{acronym}
\acro{NN}[NN]{Neural Networks}
\end{acronym}

\begin{acronym}
\acro{TL}[TL]{Tolles-Lawson}
\end{acronym}

\begin{acronym}
\acro{GNSS}[GNSS]{Global Navigation Satellite System}
\end{acronym}

\begin{acronym}
\acro{MagNav}[MagNav]{Magnetic Anomaly Navigation}
\end{acronym}

\begin{acronym}
\acro{EKF}[EKF]{extended Kalman filter}
\end{acronym}

\begin{acronym}
\acro{ESKF}[ESKF]{error-state Kalman filter}
\end{acronym}

\begin{acronym}
\acro{RMSE}[RMSE]{Root-Mean-Square Error}
\end{acronym}

\begin{acronym}
\acro{CRLB}[CRLB]{Cramér-Rao Lower Bound}
\end{acronym}

\begin{acronym}
\acro{IMA}[IMA]{Integrated Modular Avionics}
\end{acronym}

\begin{acronym}
\acro{ML}[ML]{Machine-Learning}
\end{acronym}

\begin{acronym}
\acro{IAE}[IAE]{Innovation-Based Adaptive Estimation}
\end{acronym}

\begin{acronym}
\acro{IMU}[IMU]{inertial measurement unit}
\end{acronym}

\begin{acronym}
\acro{RNP}[RNP]{Required Navigation Performance}
\end{acronym}

\begin{acronym}
\acro{ECEF}[ECEF]{Earth-centered Earth-fixed}
\end{acronym}

\begin{acronym}
\acro{FOGM}[FOGM]{First-Order Gauss-Markov}
\end{acronym}

\begin{acronym}
\acro{IGRF}[IGRF]{International Geomagnetic Reference Field}
\end{acronym}

\begin{acronym}
\acro{WGS84}[WGS84]{World Geodetic System 1984}
\end{acronym}

\begin{acronym}
\acro{INS}[INS]{inertial navigation system}
\end{acronym}

\begin{acronym}
\acro{RMSE}[RMSE]{Root Mean Square Error}
\end{acronym}

\begin{acronym}
\acro{PNT}[PNT]{positioning, navigation and timing}
\end{acronym}

\begin{acronym}
\acro{UAV}[UAV]{unmanned aerial vehicle}
\end{acronym}

\begin{acronym}
\acro{MC}[MC]{Monte Carlo}
\end{acronym}

\begin{acronym}
\acro{MEMS}[MEMS]{micro-electro-mechanical systems}
\end{acronym}

\begin{acronym}
\acro{NED}[NED]{North-East-Down}
\end{acronym}

\title{{Closed-loop vs. Open-loop Kalman Filter Architectures in Airborne Aided Inertial Navigation}}

\author{
\authorblockN{
	Antonia Hager\authorrefmark{1}\authorrefmark{2} and
	Torleiv H. Bryne\authorrefmark{2}
}
\authorblockA{\authorrefmark{1}\textit{Airbus Central Research and Technology, Taufkirchen, Germany}\\}
\authorblockA{\authorrefmark{2}\textit{Department of Engineering Cybernetics, Norwegian University of Science and Technology, Trondheim, Norway}\\}
}

\maketitle

\thispagestyle{preprintnotice}

\begin{abstract}
Closed-loop (or feedback) \acp{ESKF} with their relatives and offspring are the state-of-the-art in modern aided inertial navigation research.
Estimated \ac{INS} errors are continually fed back to the \ac{INS} to correct the nominal system state before subsequent predictions. Conversely, in safety-critical aeronautical applications, open-loop (or feedforward) systems are an undisputed standard, where the inertial mechanization is strictly decoupled to allow for operational independence and fault isolation of computing units.
We assess the performance impacts of this architectural choice beyond qualitative system-safety justifications using a standard inertial mechanization in geodetic coordinates and direct position aiding. Simulations using a variety of inertial sensor error characteristics, ranging from consumer to navigation grade systems, showcase the trade-off between smooth information fusion for high-end \acs{IMU}s using an open-loop filter and the inherent long-term stability of the closed-loop architecture.
\vspace{-0.2cm}
\end{abstract}

\section{Introduction}

Modern aerospace and robotic autonomous systems require highly robust, continuous, and reliable positioning capabilities. Although \acp{GNSS} provide absolute positioning, their vulnerability to signal attenuation, jamming, or spoofing results in a reliance on an \ac{INS} as the primary means of navigation, supported by other information sources~\cite{Engelsman2023}. \acs{INS} rely on an \ac{IMU} with accelerometers and gyroscopes to provide position, velocity, and attitude predictions via mathematical integration of kinematic equations~\cite{Groves2013,Titterton}. A well-known fundamental limitation of inertial navigation is that deterministic and stochastic sensor errors are integrated through the navigation equations, causing the estimation error to grow unbounded over time~\cite{Maybeck79, Groves2013}.

To limit \ac{INS} drift, multi-sensor data fusion is deployed within an integrated navigation architecture, utilizing complementary sensor measurements from alternative \ac{PNT} sources such as barometers, radio signals (GNSS, DME, VOR), or the magnetic field \cite{Canciani2017, Hager2025}. In the state-of-the-art academic literature, these integrated systems are predominantly formulated using a closed-loop architecture, where state corrections and sensor bias errors estimated by the filter are continuously fed back to correct the system's full state~\cite{Engelsman2023}. This architectural paradigm has further evolved into invariant and equivariant filtering frameworks, which utilize Lie-group error definitions in the body frame to achieve state-independent uncertainty propagation and guaranteed local consistency~\cite{BarrauBonnabel2018_IKF, whittaker_inertial_2017, maurer_equivalence_2025}.

Currently, there is an architectural divergence between safety-critical airborne platforms and modern autonomous unmanned systems. Commercial airliners, transport aircraft, and other large, fast-moving, passenger-carrying platforms typically utilize high-end navigation-grade \acp{IMU}~\cite[Ch.~4]{Groves2013}. Because these high-end sensors exhibit very low bias and bias instability, industrial aerospace implementations favor open-loop (feedforward) architectures~\cite{Canciani2017, Canciani2022, Kim2007}. In this configuration, the primary dead-reckoning computations run on independent hardware to meet safety regulations, while an external estimation processor models system errors without feeding corrections back into the integration loop. This isolation ensures that even if a sensor anomaly or filter divergence occurs within an auxiliary positioning filter, the primary inertial reference position remains intact.

In contrast, small-scale autonomous systems, such as \acp{UAV} or robotic platforms, operate under tighter size, weight, power, and cost constraints. These platforms are restricted to lower-grade \ac{MEMS} \acp{IMU} with more significant bias instabilities~\cite{Suvorkin2024}. To maintain the stability of the navigation solution, these systems typically use a closed-loop architecture, in which estimated errors and sensor bias errors are continuously injected back into the full state of the navigation system \cite[Ch.~14]{Groves2013}. 

Alternative \ac{PNT} solutions are used for smaller, highly dynamic, autonomous flying vehicles where navigation-grade hardware is not standard. Selecting an optimal filter architecture requires balancing the hardware constraints of lower-grade sensors against architectural isolation for platform safety. Despite textbook descriptions detailing both configurations~\cite[Ch.~6]{Maybeck79},\cite[Ch.~14]{Groves2013}, direct one-to-one comparisons that isolate the impact of this architectural choice within identical coordinate spaces remain sparse in the literature. For example, \cite{wendel_direct_2001} compared a total state (direct) and an error state (indirect) feedback filter, but did not study open-loop filters. Understanding the structural trade-offs between open- and closed-loop filters is important to determine the appropriate filter architecture for a given platform, depending on available \ac{IMU} quality and safety requirements.
\subsection{Main contributions}

We address this gap in the literature by presenting a one-to-one comparative analysis of an open- and a closed-loop navigation filter with position updates. To isolate the structural behavior of the error propagation loops, both configurations use an identical mathematical baseline: a 15-state \ac{ESKF} with navigation-frame (or, correspondingly, world-frame in robotics) estimation errors. We evaluate these feedforward and feedback implementations for various \ac{IMU} error specifications and describe the operational boundaries separating aerospace software standards from feedback-based navigation filter algorithms predominant in recent literature. 

\subsection{Outline}
\Cref{sec:Preliminaries} establishes the coordinate frames and mathematical notation used throughout this manuscript. 
\Cref{sec:filter_equations} describes the 3-D inertial navigation equations and their error-state formulations in geodetic coordinates.
\Cref{sec:open_closed_loop} details the structural differences between the closed- and the open-loop architecture.
In \Cref{sec:implementation} we present the results of our comparative simulations under diverse \ac{IMU} noise characteristics ranging from consumer to navigation-grade systems.
Finally, \cref{sec:conclusion} summarizes our core findings and presents closing remarks regarding platform-specific architectural selection.

\section{Preliminaries}
\label{sec:Preliminaries}
\begin{table}[tb]
\begin{center}
\centering
\vspace{0.2cm}
\caption{\label{tab:notation}Notation}
\begin{tabularx}{\columnwidth}{>{\centering\arraybackslash}p{2cm} | X}
        \toprule \bfseries
        Symbol & \bfseries Description \\ 
        \midrule
        \inertial, \ecef, \body, \ned, \geo  & Sub-/superscripts for inertial, ECEF-, body-, NED and geodetic coordinate frames  \\   
        $\bfx, \bfp, ...$ & Vector quantities denoted by bold lowercase letters \\
        $\bfF, \bfW, ...$ & Matrix quantities denoted by bold uppercase letters \\
        \midrule
        $\bm p$, $\bm v$, $\bm{\vartheta}$ & Position, velocity, and attitude vectors\\
        $\mu, \lambda, h$ & Latitude, longitude, and ellipsoidal altitude, \\
        &  $\mu \in [-\pi/2, \pi/2]$,   $\lambda \in (-\pi, \pi]$, $h\in \mathbb{R}^1$  \\
        %
        $\bm{R}_{nb}$ & Rotation matrix from $\{b\}$ to $\{n\}$ frame\\
        $\bm{q}_{nb}$ & Unit quaternion from $\{b\}$ to $\{n\}$ frame\\
        \hline
        $R_0$ & Equatorial radius $R_0 = \SI{6378137.0}{\meter}$ \\
        $f$, $e$ & Flattening $f = 1 / 298.257223563$ and first eccentricity $e = f(2-f)$ of the reference ellipsoid \\
        $R_M$, $R_N$ & Earth's Meridial and Normal Radius \\
        $\omega_{ie}$ & Earth's rotation rate $\omega_{ie} = \SI{7.292115e-5 }{\radian\per\second}$\\
        $\bfg^n$ & Gravitational acceleration in NED frame
        \\
        \midrule
        $\bfa_m$, $\boomega_m$ & Acceleration and angular rate measured by IMU \\
        $\boeta_{\bfa}$, $\boeta_{\boomega}$ & Angular rate and acceleration measurement noise \\
        $\bfa_{\bfb}$, $\boomega_{\bfb}$ & Accelerometer and gyroscope bias\\
        $\boeta_{\bfa \bfb}$, $\boeta_{\boomega \bfb}$ & Accelerometer and gyroscope bias errors \\        
        \midrule
        $\bfx$, $\delta \bm x$ & Nominal state, Error state \\     
        \midrule
        $\hbfR, \hbfx, \delta \hbfx, ...$ & Estimated quantities\\
        \bottomrule
    \end{tabularx}
\end{center}
\vspace{-0.7cm}
\end{table}
The navigation frame \ned~is defined as the right-handed local tangent frame with its origin aligned with the vehicle's body frame \body, and its axis pointing to geographic \ac{NED} directions.
Global position is expressed in geodetic coordinates, $\bfp^g = \begin{bmatrix} \mu & \lambda & h \end{bmatrix}^\top$. Latitude and longitude are measured in radians, and altitude is expressed in meters above the WGS-84 ellipsoid surface. 
The definitions of the variables are given in \cref{tab:notation}. Superscripts define the coordinate system in which a quantity is expressed. 
\section{Geodetic Navigation Equations}
\label{sec:filter_equations}
The nominal 3-D state of a vehicle is described by a position in geodetic coordinates, $\bfp_g$, and its velocity $\bfv_{nb}^n$ and attitude quaternion $\bfq_{nb}$  resolved in the local NED-frame,
\begin{align}
    \bfx = \begin{bmatrix}
        \bfp^{g\top} & \bfv_{nb}^{n\top} & \bfq_{nb}^{\top}
    \end{bmatrix}^\top.
    \label{eq:nominal_state}
\end{align}
The quaternion is related to the 3D rotation vector $\bovartheta$ by 
\begin{align}
    \bfq(\bovartheta) =
    \begin{bmatrix}
        \cos(\vartheta/2) \\
        \sin(\vartheta/2) \bovartheta/\vartheta
    \end{bmatrix}
    \underset{\|\bovartheta\|_2 \ll 1}{\approx} 
    \begin{bmatrix}
        1\\
        \frac{1}{2}\bovartheta \bf
    \end{bmatrix}.
    \label{eq:quat_exp_map}
\end{align}
\subsection{Nominal State Kinematics}

The kinematics of the nominal state are used in the closed-loop filter to propagate the measurement-corrected nominal state. We use the discretized inertial equations in geodetic coordinates resumed in the following.
The discretization time step is denoted by $\Delta t := \Delta t_{k-1,k} = t_k -t_{k-1}.$
The ellipsoidal radii of curvature, cf. \cref{tab:notation}, and navigation-frame transport rates are updated via:
\begin{align}
R_N &= \frac{R_0}{\sqrt{1 - e^2 \sin^2(\mu)}}, \quad R_M = \frac{R_0(1 - e^2)}{(1 - e^2 \sin^2(\mu))^{1.5}}, \\
\boomega_{ie}^n &= \begin{bmatrix} \omega_{ie} \cos(\mu) & 0 & -\omega_{ie} \sin(\mu) \end{bmatrix}^T, \\
\boomega_{en}^n &= \begin{bmatrix} \frac{v_\text{E}}{R_N + h} & -\frac{v_\text{N}}{R_M + h} & -\frac{v_\text{E} \tan(\mu)}{R_N + h} \end{bmatrix}^T,
\end{align}
where $\boomega_{in}^n = \boomega_{ie}^n + \boomega_{en}^n$.

\subsubsection*{Attitude}
The attitude quaternion $\bfq_{nb}$ is integrated using the incremental rotation quaternions $\bfq_{n_{k}}^{n_{k-1}}$ and $\bfq_{b_{k-1}}^{b_{k}}$:
\begin{align}
\bfq_{nb, \, k} &= \bfq_{n_{k}}^{n_{k-1}} \otimes 
 \bfq_{nb, \, k-1} \otimes \bfq_{b_{k-1}}^{b_{k}},
\label{eq:att_kinematics}
\end{align}
where $\bfq_{n_{k}}^{n_{k-1}} = \exp(-\Delta t \, \boomega_{in}^n)$, and $\bfq_{b_{k-1}}^{b_{k}} = \exp(\Delta t \, \boomega_{ib}^b)$.

\subsubsection*{Velocity}
The velocity is updated using the current specific force $\bfa_{eb}^n$ in the navigation frame:
\begin{align}
\label{eq:vel_kinematics}
\begin{split}
\bfa_{eb}^n &= \bfR_{nb} \bff_{ib}^b - \bfS (2\boomega_{ie}^n + \boomega_{en}^n) \bfv_{k-1} + \bfg^n_b, \\
\bfv_k &= \bfv_{k-1} + \bfa_{eb}^n \Delta t.
\end{split}
\end{align}
%
The navigation-frame gravity $\bfg_b^n = \bfg^n_b(\mu, \lambda)$ is calculated as in \cite[Ch.~2.4.7]{Groves2013}.

\subsubsection*{Position}
We use a second-order trapezoidal integration scheme to update the geodetic position vector $p_i^g$:
\begin{align}
\mu_k &= \mu_{k-1} + \frac{\Delta t}{2} \frac{(v_{n, k} + v_{n, k-1})}{R_M + h_{k-1}}, \\
\lambda_k &= \lambda_{k-1} + \frac{\Delta t}{2} \frac{(v_{e, k} + v_{e, k-1})}{(R_N + h_{k-1})\cos(\mu_{k-1})}, \\
h_k &= h_{k-1} - \frac{\Delta t}{2} (v_{d, k} + v_{d, k-1}).
\label{eq:pos_kinematics}
\end{align}

\subsection{Error State Kinematics}

The error state is defined as
\begin{equation}
   \delta \bfx \triangleq  \begin{bmatrix}
        \delta \bfp^{g\top} & \delta \bfv_{nb}^{n\top} & \delta \bm{\vartheta}_{nb}^{n\top} &  \bfa^{b\top}_{\bfb} & \boomega^{b\top}_{\bfb}
    \end{bmatrix}^\top,
\label{eq:error_state_total}
\end{equation}
where the attitude (error) is now represented as a rotation vector $\delta \bovartheta_{nb}^{n}$
instead of a quaternion $\bfq_{nb} ( \bovartheta_{nb}^{n^\top} )$.

The error-state kinematics are the linearized global inertial navigation equations in geodetic coordinates, given by:
\begin{equation}
    \delta \dot{\bfx} = \bfF \delta {\bfx} + \bfG \bfw \label{eq:err_kinematics}
\end{equation}
where
\begin{equation*}
\begin{aligned}[c]
\bfF &= \begin{bmatrix} 
\bfF_{pp} & \bfF_{pv} & \bm{0}_{3\times3} & \bm{0}_{3\times3} & \bm{0}_{3\times3} \\ 
\bfF_{vp} & \bfF_{vv} & -\bfS(\bfR_{nb} \bff_{ib}^b) & \bfR_{nb} & \bm{0}_{3\times3} \\ 
\bfF_{\vartheta p} & \bfF_{\vartheta v} & \bfF_{\vartheta\vartheta} & \bm{0}_{3\times3} & -\bfR_{nb} \\ 
\bm{0}_{3\times3} & \bm{0}_{3\times3} & \bm{0}_{3\times3} & -p_{\bfa \bfb}\bm{I}_{3} & \bm{0}_{3\times3} \\ 
\bm{0}_{3\times3} & \bm{0}_{3\times3} & \bm{0}_{3\times3} & \bm{0}_{3\times3} & -p_{\boomega \bfb}\bm{I}_{3} 
\end{bmatrix}
\\
\bfG &=
  \begin{bmatrix}
        \bfzero_3 & \bfzero_3  & \bfzero_3 & \bfzero_3 \\
        \bfR_{nb} & \bfzero_3  & \bfzero_3 & \bfzero_3 \\
        \bfzero_3 & -\bfR_{nb} & \bfzero_3 & \bfzero_3 \\
        \bfzero_3 & \bfzero_3  & \bfI_3    & \bfzero_3 \\
        \bfzero_3 & \bfzero_3  & \bfzero_3 & \bfI_3
    \end{bmatrix},
\end{aligned}
\end{equation*}
with $\bfF_{pv} = \text{diag}\left( \frac{1}{R_M+h}, \frac{1}{(R_N+h) \cos(\mu)}, -1 \right)$. The remaining sub-matrices are summarized in \cref{tab:pinson_sub_matrices}.

\begin{table*}[t]
\centering
\caption{Position, velocity, and attitude sub-matrices of the linearized error state kinematics}
\label{tab:pinson_sub_matrices}
\scriptsize
\vspace{-6pt}
\begin{align*}
\bfF_{vp} &= \begin{bmatrix} -v_\text{E}\left(2\omega_{ie}\cos\mu + \frac{v_\text{E}}{(R_N+h)\cos^2\mu}\right) & 0 & \frac{v_\text{E}^2\tan\mu - v_\text{N} v_\text{D}}{(R_M+h)^2} \\ 2\omega_{ie}(v_\text{N}\cos\mu - v_\text{D}\sin\mu) + \frac{v_\text{N} v_\text{E}}{(R_N+h)\cos^2\mu} & 0 & -v_\text{E}\frac{v_\text{N}\tan\mu + v_\text{D}}{(R_N+h)^2} \\ 2\omega_{ie} v_\text{E} \sin\mu & 0 & \frac{v_\text{N}^2}{(R_M+h)^2} + \frac{v_\text{E}^2}{(R_N+h)^2} \end{bmatrix} & 
\bfF_{\vartheta p} &= \begin{bmatrix} -\omega_{ie}\sin\mu & 0 & -\frac{v_\text{E}^2}{(R_N+h)^2} \\ 0 & 0 & \frac{v_\text{N}}{(R_M+h)^2} \\ -\omega_{ie}\cos\mu - \frac{v_\text{E}}{(R_N+h)\cos^2\mu} & 0 & \frac{v_\text{E}\tan\mu}{(R_N+h)^2} \end{bmatrix}
\end{align*} \\
%
\begin{align*}
\bfF_{vv} &= \begin{bmatrix} \frac{v_\text{D}}{R_M+h} & -2\left(\omega_{ie}\sin\mu + \frac{v_\text{E}\tan\mu}{R_N+h}\right) & \frac{v_\text{N}}{R_M+h} \\ 2\omega_{ie}\sin\mu + \frac{v_\text{E}\tan\mu}{R_N+h} & \frac{v_\text{N}\tan\mu + v_\text{D}}{R_N+h} & 2\omega_{ie}\cos\mu + \frac{v_\text{E}}{R_N+h} \\ -\frac{2v_\text{N}}{R_M+h} & -2\left(\omega_{ie}\cos\mu + \frac{v_\text{E}}{R_N+h}\right) & 0 \end{bmatrix}
&\bfF_{\vartheta v} &= \begin{bmatrix} 0 & \frac{1}{R_M+h} & 0 \\ -\frac{1}{R_M+h} & 0 & 0 \\ 0 & -\frac{\tan\mu}{R_N+h} & 0 \end{bmatrix}
\end{align*}
\begin{align*}
\bfF_{pp} = \begin{bmatrix} 0 & 0 & -\frac{v_\text{N}}{(R_M+h)^2} \\ \frac{v_\text{E} \tan\mu}{(R_N+h) \cos\mu} & 0 & -\frac{v_\text{E}}{(R_N+h)^2 \cos\mu} \\ 0 & 0 & 0 \end{bmatrix}  & & & & 
\bfF_{\vartheta\vartheta} = \begin{bmatrix} 0 & -\omega_{ie}\sin\mu - \frac{v_\text{E}\tan\mu}{R_N+h} & \frac{v_\text{N}}{R_M+h} \\ \omega_{ie}\sin\mu + \frac{v_\text{E}\tan\mu}{R_N+h} & 0 & \omega_{ie}\cos\mu + \frac{v_\text{E}}{R_N+h} \\ -\frac{v_\text{N}}{R_M+h} & -\omega_{ie}\cos\mu - \frac{v_\text{E}}{R_N+h} & 0 \end{bmatrix}
\end{align*} 
\vspace{-0.6cm}
\end{table*}

\subsubsection*{IMU Bias Dynamics}
The accelerometer and gyroscope biases that are part of the error state are modeled with a discrete-time first-order Gauss-Markov process:
\begin{align}
\bfa_{\bfb, k}^b &=\bfa_{\bfb, k-1}^b e^{-\Delta t \, p_{\bfa \bfb}}, \quad \boomega_{\bfb, k}^b = \boomega_{\bfb, k-1}^b e^{-\Delta t \, p_{\boomega \bfb} },
\label{eq:bias_kinematics}
\end{align}
with inverse time constants $p_{\star}= 1/\tau_\star$ and bias process driving noises $\boeta^b_{\bfa \bfb} \sim \mathcal{N} ( \bm 0, \bosigma_{\bfa \bfb})$, $\boeta^b_{\boomega \bfb} \sim \mathcal{N} ( \bm 0, \bosigma_{\boomega \bfb})$.

\subsection{Measurement Updates}
\label{sec:measurement_updates}
We employ a loosely coupled filter where a position measurement is used for aiding, e.g. derived from a \ac{GNSS} signal or other positioning means. The filter receives it as a position measurement in the navigation frame.
The error state $\delta \bfx$ gets updated sequentially for every new measurement using the standard Kalman filter update equations
\begin{align}
    \bfW_k &= \covP_{k|k-1} \bfH_k^\top (
    \bfH_k \covP_{k|k-1} \bfH^\top_k + \covR_{k}
    )^{-1} \\
    \delta \hbfx_k &= \bfW_k ( \bfy - \bfh(\hbfx_{k|k-1}) ) \\
    \covP_{k} &= (\bfI_{15} - \bfW \bfH_k) \covP_{k|k-1}.
\end{align}
%
As the filter's errors are defined in the navigation frame, the measurement equation and corresponding Jacobian become
\begin{align}
\begin{split}
    \bfy &= \bfh(\bfx) + \bfw_\mathrm{p} = \bfp_{nb}^n + \bfw_\mathrm{p}  \\ 
    \bfH &=  \frac{\partial \bfx}{\partial \delta \bfx} = \begin{bmatrix}
        \bfI_{3 \times 3} & \bfzero_{3 \times 3} & \bfzero_{3 \times 3} & \bfzero_{3 \times 3} & \bfzero_{3 \times 3} 
    \end{bmatrix}.
\end{split} 
\label{eq:position_measurement}
\end{align}
Due to the linearity of the position error. The position measurement is subject to white Gaussian noise $\bfw \sim \mathcal{N}(0, \bosigma_{\bfp}^2)$.

\section{Open- and Closed-loop Architectures}
Both architectures share the same dynamic model for the error state and the same measurement updates.
The error state covariance update is the standard \ac{ESKF} covariance formula
\begin{align}
    \covP_{k|k-1} &= 
    \boPhi_{k-1} \covP_{k-1} 
    \boPhi_{k-1}^\top + 
    \covQ_d,
    \label{eq:P_pred}
\end{align}
where the discrete transition matrix $\boPhi_{k-1}$ and the process noise covariance $\covQ_d$ are obtained by discretizing system~\cref{eq:err_kinematics} 
and the spectral density matrix $\covQ_c = \diag(\bosigma_{\bfF}^2, \bosigma_{\boomega}^2, \bosigma_{\bfF\bfb}^2, \bosigma_{\boomega\bfb}^2)$ using Van Loan's formula \cite{van_loan_computing_1978}. The $\covQ_c$ matrix is tuned according to the respective IMU specifications.

The open and closed-loop architectures differ in how corrections of the error state are applied to the nominal state, and consecutively in their nominal state evolution.
Since the correction of the nominal state relies on the linearity approximation, the differences between feedback and open-loop corrections impact the applicability of one or the other architecture for different use cases.

\subsection{Open-loop corrections}
\label{sec:open_closed_loop}
The filter is based on a drifting dead-reckoning solution 
$\bfx_{\text{INS}} = [\bfp_{\text{INS}}^{g\top},\, \bfv_{nb,\text{INS}}^{n\top},\, \bfq_{nb,\text{INS}}^{\top} ]^\top$,
that is maintained outside the state estimation loop. 
The error state, $\delta \bfx$, is initialized to zero and propagated using 
\begin{equation}
       \delta \hbfx _{k|k-1} = \boPhi_{k-1} \cdot \delta \hbfx_{k-1}.\label{eq:open_loop_error_state_prop}
\end{equation}
It grows unbounded over time without a reset to zero, as opposed to the repeating reset of the error state in closed-loop filtering. Its covariance stays contained by the measurement updates. The final nominal state estimate is calculated as a composition of the INS-state and the state error:
\begin{align}
    \hbfx = \bfx_{\text{INS}} \oplus \delta \hbfx.
\end{align}
Hence, in the open-loop architecture, the kinematically integrated navigation states ($\bfp_{\text{INS}}^{g}, \bfv_{nb,\text{INS}}^{n},\bfq_{nb,\text{INS}}$) are corrected externally for output or filtering, but the internal error state of the filter remains without reset. The bias corrections are directly propagated in the error state estimate during \cref{eq:open_loop_error_state_prop} via the rotation matrices in the upper right part of the $\bfF$ matrix. 

The state correction vector maps as:
\begin{align}
\hbfp_k^g &= 
\begin{bmatrix}
\mu_{\text{INS},k|k-1} + \delta\hat{\mu}_{k}\\
\lambda_{\text{INS},k|k-1} + \delta\hat{\lambda}_{k} \\
h_{\text{INS},k|k-1} + \delta \hat{h}_{k}
\end{bmatrix} \\
\hbfv_{nb,k-1}^n &= \bfv_{nb,\text{INS},k|k-1}^{n} + \delta\hbfv_{nb,k}^n \\
\hbfq_{nb, k} &= \hbfq({\delta\bm{\vartheta}_k}) \otimes \bfq_{nb,\text{INS}, k|k-1},
\end{align}
where \cref{eq:quat_exp_map} is used to convert the navigation-frame (world) attitude error vector to an error quaternion.
Notice that the sensor biases remain isolated ($\bfa_{\bfb,k-1}^b = \bfa_{\bfb,k|k-1}^{b}$). 

The nominal estimate is the final output of the filter. It is further used in \cref{eq:position_measurement} to obtain the predicted measurement for the update and to propagate the error state using \cref{eq:open_loop_error_state_prop}, but it is not propagated itself.
With this, the filter continuously maintains three states: the INS solution, the error state, and the nominal state for the output.

\subsection{Closed-loop corrections}

The nominal state is propagated according to \cref{eq:att_kinematics}, \cref{eq:vel_kinematics} and \cref{eq:pos_kinematics} using the uncorrected inertial measurements $\bff^b_m$ and
$\boomega^b_m$. The error state covariance propagtion via \cref{eq:err_kinematics} uses the bias-corrected measurements $\hbff^b_{ib}$ and $\hboomega_{ib}^b$, and the estimated rotation $\hbfR_{nb}$.
Using the uncorrected measurements for the propagation step avoids double-integration of the biases that remain solely in the error state. 

The measurement update in \cref{sec:measurement_updates} is performed on the error state. Then, in the closed-loop architecture, only the estimated kinematic errors are fed back directly to update the total system state. Once injected, the position, velocity, and attitude error states are reset to zero, resulting in \cref{eq:open_loop_error_state_prop} evaluating to zero. The state correction vector is formulated as:
\begin{align}
\hbfp_k^g &= \begin{bmatrix}
\hat{\mu}_{k|k-1} + \delta\hat{\mu}_k \\
\hat{\lambda}_{k|k-1} + \delta\hat{\lambda}_k \\
\hat{h}_{k|k-1} + \delta \hat{h}_k
\end{bmatrix} \\
\hbfv_{nb,k-1}^n &= \hbfv_{nb,k|k-1}^{n} + \delta\hbfv_{nb,k}^n \\
\hbfq_{nb, k} &= \hbfq({\delta\bm{\vartheta}_k}) \otimes \hbfq_{nb, k|k-1}^{}.
\end{align}

The bias dynamics can evolve either in the INS or in the error state. Using the latter, the parts of \cref{eq:open_loop_error_state_prop} related to bias are non-zero. Using the former, the bias correction is included in the propagation of the nominal state directly by the correction of the position error. The nominal and error states of position, velocity, and attitude are represented in the same coordinate frame.

\section{Simulation Setup and Results}
\label{sec:implementation}
\begin{figure}[tb]
    \centering
    \includegraphics[width=0.95\columnwidth]{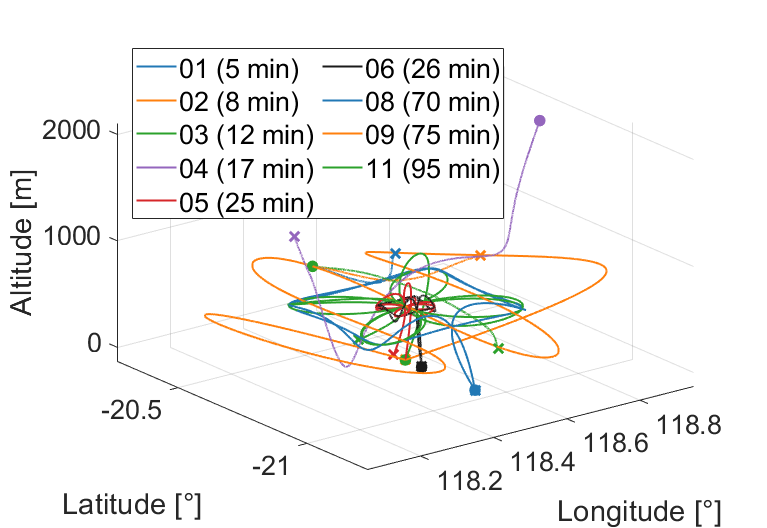}
    \caption[  ]
    {\small 3-D profiles of the dynamic trajectories; straight paths \{07,10\} are not shown.} 
    \label{fig:trajectories}
    \vspace{-0.3cm}
\end{figure}

The two filters are evaluated across diverse trajectories ranging from \SI{5}{} to \SI{95}{\minute} to assess short- and long-term performance.
The dataset comprises highly dynamic flight profiles with an average velocity of \SI{100}{\kmh}, illustrated in \cref{fig:trajectories}, and two straight flights along a geodetic path, lasting \SI{30}{} and \SI{90}{\minute} at a cruise velocity of \SI{350}{\kmh}, with a periodic attitude excitation of $\pm \SI{5}{\degree}$ with a \SI{100}{\second} period to maintain system observability.

Performance is compared over 50 \ac{MC} runs per configuration, sweeping across consumer-, tactical\nobreak, and navigation-grade sensor noise profiles as detailed in \cref{tab:imu_specs}. Both \ac{IMU} data and absolute position measurements in geographical coordinates are sampled at \SI{20}{\hertz}. Position observations are corrupted by zero-mean white Gaussian noise with a standard deviation of \SI{1}{\meter}, projected into latitude and longitude components. The initial state covariance for the accelerometer bias is identical across all sensor grades. A run is interrupted when the position error exceeds \SI{500}{\meter} or the filter becomes highly inconsistent. To examine robustness against unmodeled deterministic initialization errors, a separate set of trials introduces a constant initial acceleration bias offset of \SI{0.5}{\meter\per\second^2} across all grades. While this is not representative of calibrated navigation-grade instruments, it isolates the filters' structural convergence properties.

\begin{table}[tb]
\begin{center}
\centering
\caption{\label{tab:imu_specs}IMU noise parameters for consumer, tactical, and navigation sensor grades.}
\footnotesize
    \begin{tabularx}{\columnwidth}{p{2.7cm}  p{1.6cm} p{1.8cm} p{1.4cm}}
        \toprule \bfseries
        IMU Parameter & Consumer & Tactical & Navigation\\ 
        \midrule
        VRW $\left[\unit{\meter\per{\second\tothe{\text{\sfrac32}}}}\right]$
        & $1.31 \times 10^{-2}$
        & $3.3\times 10^{-4}$  & $10^{-12}$                  \\
        Acc bias $\sigma_{\bfa \bfb}$ $\left[\unit{\meter\per{\second\tothe{2}}}\right]$  & $1.15 \times 10^{-1}$ & $9.81\times 10^{-3}$    & $10^{-5}$ \\
        Acc bias $p_{\bfa \bfb}$ $\left[\unit{\second}\right]$  & 300 & 3600    & 3600 \\
        \midrule 
        ARW $\left[\unit{\rad\per{\second\tothe{\text{\sfrac12}}}}\right]$ & $2.30 \times 10^{-3}$ & $3.6361 \times 10^{-5}$    & $10^{-7}$ \\
        Gyro bias $\sigma_{\boomega \bfb}$ $\left[\unit{\rad\per\second}\right]$ & $9.81 \times 10^{-3}$ & $4.8481 \times 10^{-6}$    & $10^{-8}$ \\
        Gyro bias $p_{\boomega \bfb}$ $\left[\unit{\second}\right]$ & 300 & 3600    & 3600 \\
        \bottomrule
    \end{tabularx}
\end{center}
\vspace{-0.7cm}
\end{table}%

\begin{figure*}[tb]
    \centering
    \includegraphics[width=1.7\columnwidth]{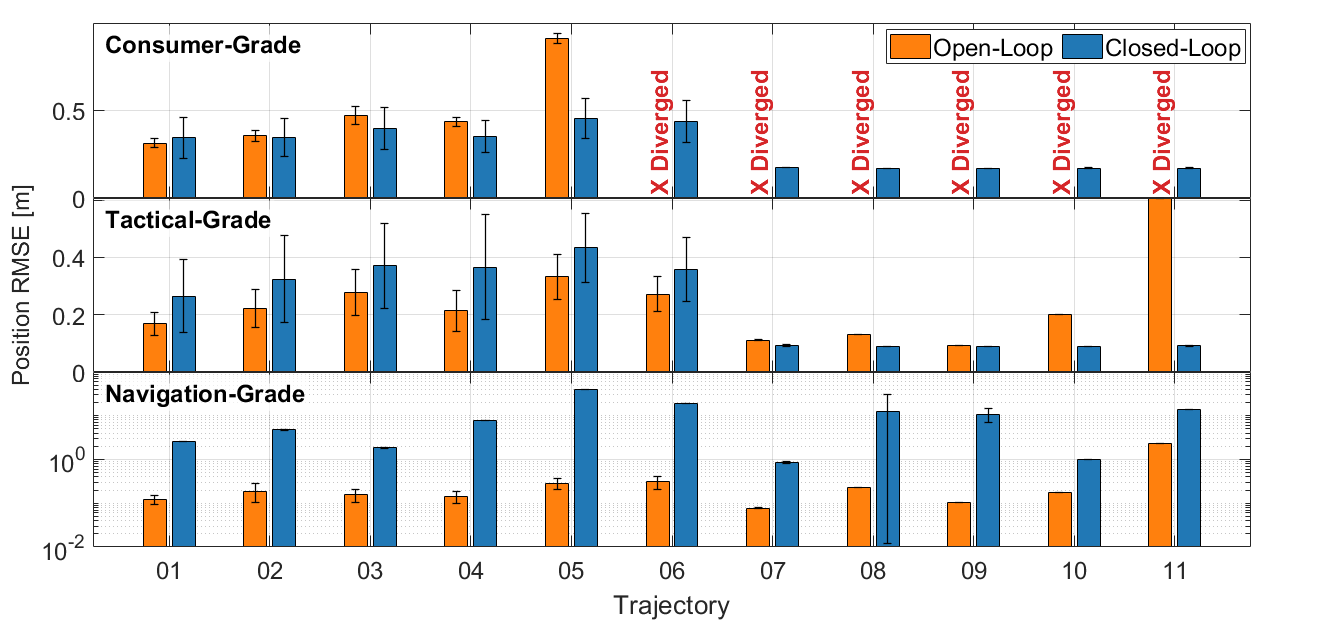}
    \caption[  ]
    {\small  Statistical Monte Carlo performance results showing total position RMSE and standard deviation per trajectory across \ac{IMU} grades, without initial accelerometer bias offset.} 
    \label{fig:mc_results}
    \vspace{-0.6cm}
\end{figure*}
\begin{figure}[tb]
    \centering
    \includegraphics[width=0.9\columnwidth]{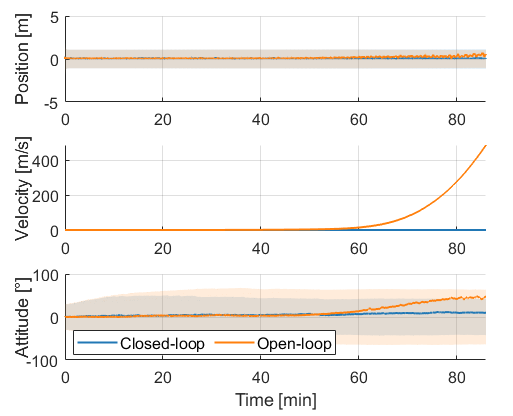}
    \caption[  ]
    {\small Position, Velocity, and attitude errors and respective $3\sigma$ uncertainty for a 90-minute straight flight (Trajectory 10) comparing the closed- and open-loop filters using a tactical grade \ac{IMU}.} 
    \label{fig:errors}
    \vspace{-0.75cm}
\end{figure}

An overview of the resulting position errors under zero initial bias is shown in \cref{fig:mc_results}.
A performance crossover between the two architectures is apparent. When \ac{INS} drift is low -- this includes both higher-grade \ac{IMU} and shorter time horizons -- the open-loop filter systematically outperforms its closed-loop counterpart. 
In our simulation, the boundary where open-loop filter performance degrades compared to closed-loop filtering is between the shorter (Trajectories 01-06 are $< \SI{30}{\min}$), and longer (Trajectories 07-11 are $> \SI{30}{\min}$) flight path simulations using tactical-grade \ac{IMU}.
This short-term stability of the open-loop system stems from the structural decoupling of the error state propagation. Noise entering the position updates cannot directly couple into the inertial integration.
Conversely, for lower-grade \acp{IMU}, the open-loop trajectory begins diverging within a few minutes across most profiles, and eventually for all grades at a long enough time horizon, leading to growing position errors towards the end of the path, as exemplarily shown in \cref{fig:errors} for one flight path. The error injection in the closed-loop filter mitigates this by avoiding the error of the nominal state from growing rapidly from the start.
However, this feedback loop penalizes navigation-grade sensors over short horizons: The feedback filter exhibits statistical inconsistency and minor bias wandering, as the injection of noisy position updates can perturb the narrow inertial integration predictions.

Because both filters evaluate their state transition matrices using the corrected state estimate, this performance difference is not driven by a linearization discrepancy as such, but by the fact that the small-error assumption underlying the linearization of the state dynamics gets violated when the drift accumulates in the open-loop filter. The feedback filter, in contrast, avoids error growth and operates within the linear regime. It sacrifices high-frequency decoupling to continuously mitigate nominal state error growth, leading to long-term bounded stability at the expense of short-term noise amplification.
Notably, dynamic flight profiles amplify integration errors, impacting particularly the open-loop architecture, as evidenced by the comparably larger position errors in Trajectories 03 and 05 across consumer and tactical grades in \cref{fig:mc_results}.

\begin{figure}[tb]
    \centering
    \includegraphics[width=0.9\columnwidth]{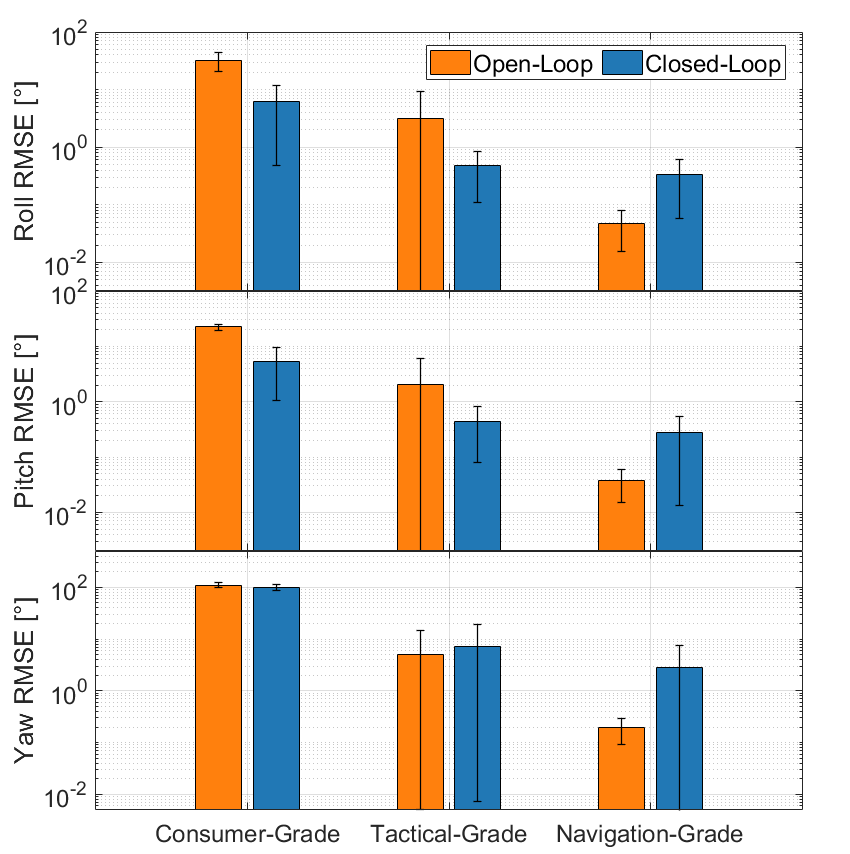}
    \caption[  ]
    {\small Attitude RMSE averaged over all trajectories and \ac{IMU} grades, without initial accelerometer bias offset.} 
    \label{fig:att_err_mc}
    \vspace{-0.8cm}
\end{figure}

\cref{fig:att_err_mc} compares the averaged attitude errors across all trajectories. The closed-loop filter results in better roll and pitch estimates across consumer and tactical \acp{IMU}. The yaw estimate is, however, more sensitive to disturbances from noisy updates due to the low observability of heading, confirming that feedforward architectures can better preserve heading integrity with good-enough sensors.

\begin{figure}[tb]
    \centering
    \includegraphics[width=0.9\columnwidth]{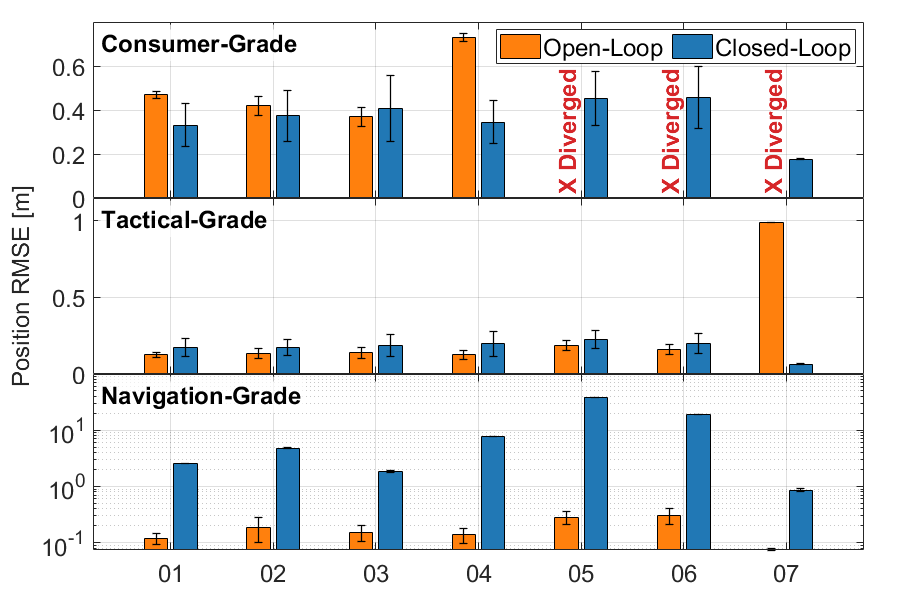}
    \caption[  ]
    {\small Total position RMSE and standard deviation per trajectory for the shorter trajectories with initial accelerometer bias offset across \ac{IMU} grades.} 
    \label{fig:mc_results_const_bias}
    \vspace{-0.7cm}
\end{figure}

Introducing an unknown constant initial bias offset emphasizes these architectural trade-offs. 
\begin{figure}[tb]
    \centering
    \includegraphics[width=0.9\columnwidth]{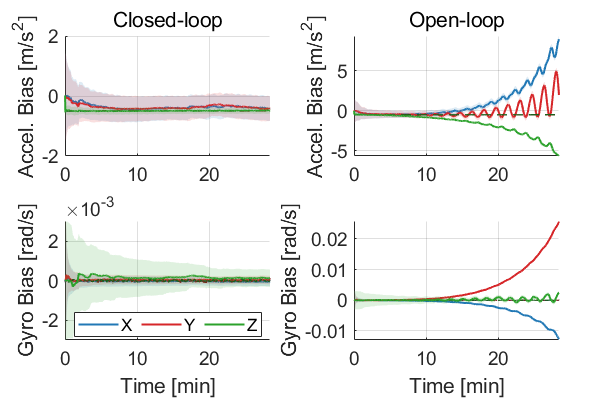}
    \caption[  ]
    {\small Accelerometer and Gyroscope bias estimates with $3\sigma$ uncertainty for a \SI{30}{\minute} straight  trajectory (07), comparing tactical-grade \ac{IMU} without and with a constant initial accelerometer bias.} 
    \label{fig:biases}
    \vspace{-0.4cm}
\end{figure}
As illustrated in \cref{fig:biases}, the unmodeled initialization error triggers an instability in the open-loop filter. Because the feedforward architecture lacks a reset path, it maintains false short-term stability until amplifying the uncorrected offset into divergence. 
This filter failure appears early in our simulations, which is why long horizons (Trajectories 08--11) are entirely omitted from the results in \cref{fig:mc_results_const_bias}. In contrast, the closed-loop filter directly isolates the initial offset, allows the state estimates to converge smoothly, and maintains bounded estimates.

The convergence behavior shows the well-known dependence on the observability of the trajectory profile. For straight flights, even the closed-loop filter takes longer to resolve the accelerometer bias due to unobservable directions, whereas dynamic maneuvers accelerate state observability. Conversely, the shorter dynamic trajectories delay the visible symptoms of the initial offset in the open-loop filter. 
\vspace{-0.2cm}

\section{Conclusion}
\label{sec:conclusion}
\vspace{-0.08cm}

This paper presented a systematic comparison of open-loop (feedforward) and closed-loop (feedback) error-state Kalman filtering architectures for airborne navigation, comparing their performance across various \ac{IMU} grades, trajectory profiles, and with initialization constraints. 
Although textbooks describe both architectures conceptually, our results provide a foundation for architecture selection based on a trade-off in error propagation. We demonstrated how open-loop systems preserve the measurement filtering and heading integrity of high-end hardware by isolating the inertial integration from measurement noise, whereas closed-loop architectures are mandatory to prevent divergence when using lower-grade sensors.

Beyond validating these known sensor boundaries, our analysis shows how structural coupling effects are sensitive to vehicle dynamics and initialization errors. Introducing an unknown initial bias reduces the stable runtime of the open-loop filter. Furthermore, when using navigation-grade \acp{IMU}, closed-loop feedback can provide a perturbation, causing minor bias wandering and degrading heading accuracy. 

For constrained autonomous platforms with lower-grade hardware, closed-loop filtering is necessary. Conversely, for safety-critical platforms equipped with high-end sensors, there are good reasons to preserve open-loop architectures not only to enforce fault isolation, but to maximize short-to-middle-term estimation accuracy. Long-term stability can also be obtained by occasional resets.
Although the extremes of the hardware spectrum unambiguously lead to the respective choices, this work provides a framework for navigation architects evaluating ambiguous design boundaries where the optimal choice is not immediately obvious. 
\vspace{-0.1cm}

\section*{Acknowledgment}
We would like to thank Philip Lillelund for valuable discussions on the open-loop implementation, and Mathieu Brunot and Tor Arne Johansen for proof-reading. We acknowledge the use of Google Gemini and Claude in editing this manuscript and programming the simulation.
\vspace{-0.1cm}
\bibliographystyle{ieeetr} 
\bibliography{refs} 

@book{Groves2013,
	edition = {2},
	title = {Principles of {GNSS}, {Intertial}, and {Multisensor} {Integrated} {Navigation} {Systems}},
	isbn = {978-1-60807-005-3},
	publisher = {Artech House},
	author = {Groves, Paul D.},
	year = {2013},
}

@book{Titterton,
	title = {Strapdown {Inertial} {Navigation} {Technology}},
	author = {Titterton, David and Weston, John},
	year = {2004},
    publisher = {IET}
}

@article{Canciani2017,
	title = {Airborne {Magnetic} {Anomaly} {Navigation}},
	volume = {53},
	issn = {0018-9251},
	url = {http://ieeexplore.ieee.org/document/7808987/},
	doi = {10.1109/TAES.2017.2649238},
	language = {en},
	number = {1},
	urldate = {2023-06-07},
	journal = {IEEE Transactions on Aerospace and Electronic Systems},
	author = {Canciani, Aaron and Raquet, John},
	month = feb,
	year = {2017},
	pages = {67--80},
}

@article{Canciani2022,
	title = {Magnetic {Navigation} on an {F}-16 {Aircraft} {Using} {Online} {Calibration}},
	volume = {58},
	issn = {0018-9251, 1557-9603, 2371-9877},
	url = {https://ieeexplore.ieee.org/document/9506809/},
	doi = {10.1109/TAES.2021.3101567},
	language = {en},
	number = {1},
	urldate = {2023-08-25},
	journal = {IEEE Transactions on Aerospace and Electronic Systems},
	author = {Canciani, Aaron J.},
	month = feb,
	year = {2022},
	pages = {420--434},
}

@incollection{Maybeck79,
	author = {Maybeck, Peter},
	booktitle = {Mathematics in Science and Engineering},
	publisher = {Academic Press},
	title = {Stochastic Models, Estimation, and Control, Volume 1},
	volume = {141},
	year = {1979}}

@article{BarrauBonnabel2018_IKF,
	author = {Axel Barrau and Silv\`{e}re Bonnabel},
	doi = {10.1146/annurev- control- 060117- 105010},
	journal = {Annual Review of Control, Robotics, and Autonomous Systems},
	pages = {237---257},
	title = {Invariant Kalman Filtering},
	volume = {1},
	year = {2018},
}

@article{van_loan_computing_1978,
	title = {Computing integrals involving the matrix exponential},
	volume = {23},
	copyright = {https://ieeexplore.ieee.org/Xplorehelp/downloads/license-information/IEEE.html},
	language = {en},
	number = {3},
	urldate = {2024-04-23},
	journal = {IEEE Transactions on Automatic Control},
	author = {Van Loan, C.},
	month = jun,
	year = {1978},
	pages = {395--404},
}

@INPROCEEDINGS{Hager2025,
  author={Hager, Antonia and Bryne, Torleiv H. and Olsen, Nils and Krauser, Jasper Simon and Johansen, Tor A.},
  booktitle={2025 European Control Conference (ECC)}, 
  title={Magnetic Anomaly Navigation with a Lie-Group Error-State Kalman Filter}, 
  year={2025},
  month={Oct.},
  volume={},
  number={},
  pages={2816-2823},
  doi={10.23919/ECC65951.2025.11187153},
  address = {Thessaloniki, Greece}
}

@ARTICLE{maurer_equivalence_2025,
  author={Maurer, Finn G. and Schmidt-Didlaukies, Henrik M. and Basso, Erlend A. and Bryne, Torleiv H.},
  journal={IEEE Transactions on Automatic Control}, 
  title={Equivalence of Left- and Right-Invariant Extended Kalman Filters on Matrix Lie Groups}, 
  year={2026},
  volume={},
  number={},
  pages={1-8},
  doi={10.1109/TAC.2026.3690028}}

@ARTICLE{Engelsman2023,
  author={Engelsman, Daniel and Klein, Itzik},
  journal={IEEE Transactions on Instrumentation and Measurement}, 
  title={Information-Aided Inertial Navigation: A Review}, 
  year={2023},
  volume={72},
  number={},
  pages={1-18},
  doi={10.1109/TIM.2023.3303496}}

@inproceedings{whittaker_inertial_2017,
	address = {Grapevine, Texas},
	title = {Inertial {Navigation} {Employing} {Common} {Frame} {Error} {Representations}},
	isbn = {978-1-62410-450-3},
	url = {https://arc.aiaa.org/doi/10.2514/6.2017-1031},
	doi = {10.2514/6.2017-1031},
	language = {en},
	urldate = {2026-05-19},
	booktitle = {{AIAA} {Guidance}, {Navigation}, and {Control} {Conference}},
	publisher = {American Institute of Aeronautics and Astronautics},
	author = {Whittaker, Matthew and Crassidis, John L.},
	month = jan,
	year = {2017},
}

@inproceedings{wendel_direct_2001,
  title={Direct Kalman filtering of GPS/INS for aerospace applications},
  author={Wendel, Jan and Schlaile, Christian and Trommer, Gert F},
  booktitle={International Symposium on Kinematic Systems in Geodesy, Geomatics and Navigation (KIS2001)},
  year={2001}
}

@article{Kim2007,
title = {FEEDFORWARD UNSCENTED KALMAN FILTER FOR INS/GPS TIGHTLY COUPLED INTEGRATION SYSTEM},
journal = {IFAC Proceedings Volumes},
volume = {40},
number = {7},
pages = {545-550},
year = {2007},
note = {17th IFAC Symposium on Automatic Control in Aerospace},
issn = {1474-6670},
doi = {https://doi.org/10.3182/20070625-5-FR-2916.00093},
url = {https://www.sciencedirect.com/science/article/pii/S1474667015332997},
author = {Kwangjin Kim and Chan Gook Park},
}

@Article{Suvorkin2024,
AUTHOR = {Suvorkin, Vladimir and Garcia-Fernandez, Miquel and González-Casado, Guillermo and Li, Mowen and Rovira-Garcia, Adria},
TITLE = {Assessment of Noise of MEMS IMU Sensors of Different Grades for GNSS/IMU Navigation},
JOURNAL = {Sensors},
VOLUME = {24},
YEAR = {2024},
NUMBER = {6},
ARTICLE-NUMBER = {1953},
URL = {https://www.mdpi.com/1424-8220/24/6/1953},
PubMedID = {38544217},
ISSN = {1424-8220},
DOI = {10.3390/s24061953}
}
\vspace{12pt}

\end{document}